\documentclass{article}

\PassOptionsToPackage{numbers, square}{natbib}

\usepackage[final]{neurips_2022}

\usepackage{macros}

\usepackage[utf8]{inputenc} 
\usepackage[T1]{fontenc}    
\usepackage{hyperref}       
\usepackage{url}            
\usepackage{booktabs}       
\usepackage{amsfonts}       
\usepackage{nicefrac}       
\usepackage{microtype}      
\usepackage{xcolor}         
\usepackage{mathrsfs}

\usepackage{algorithm}
\usepackage[noend]{algpseudocode}

\newcommand{\mm}[0]{\mathbf{m}}
\newcommand{\hh}[0]{\mathbf{h}}

\newcommand{\encoder}[0]{\textit{Snapshot Encoder}}
\newcommand{\gnn}[0]{\textsc{Gnn}}
\newcommand{\model}[0]{\textsc{AnoMulY}}
\newcommand{\V}[0]{\mathcal{V}}
\newcommand{\E}[0]{\mathcal{E}}
\newcommand{\LL}[0]{\mathcal{L}}
\newcommand{\GG}[0]{\mathcal{G}}
\newcommand{\X}[0]{\mathcal{X}}

\newcommand{\W}[0]{\mathbf{W}}

\newcommand{\squishlist}{
 \begin{list}{$\bullet$}
  { \setlength{\itemsep}{0pt}
     \setlength{\parsep}{3pt}
     \setlength{\topsep}{3pt}
     \setlength{\partopsep}{0pt}
     \setlength{\leftmargin}{1.5em}
     \setlength{\labelwidth}{1em}
     \setlength{\labelsep}{0.5em} } }

\newcommand{\squishlisttwo}{
 \begin{list}{$\bullet$}
  { \setlength{\itemsep}{0pt}
    \setlength{\parsep}{0pt}
    \setlength{	opsep}{0pt}
    \setlength{\partopsep}{0pt}
    \setlength{\leftmargin}{2em}
    \setlength{\labelwidth}{1.5em}
    \setlength{\labelsep}{0.5em} } }

\newcommand{\squishend}{
  \end{list}  }

\title{Anomaly Detection in Multiplex Dynamic Networks: from Blockchain Security to Brain Disease Prediction}

\author{%
  Ali Behrouz \\
  Department of Computer Science\\
  University of British Columbia\\
  Vancouver, BC, Canada \\
  \texttt{alibez@cs.ubc.ca} \\
  \And
  Margo Seltzer \\
  Department of Computer Science\\
  University of British Columbia\\
  Vancouver, BC, Canada \\
  \texttt{mseltzer@cs.ubc.ca} \\
}

\begin{document}

\maketitle

\begin{abstract}
The problem of identifying anomalies in dynamic networks is a fundamental task with a wide range of applications. However, it raises critical challenges due to the complex nature of anomalies,  lack of ground truth knowledge, and complex and dynamic interactions in the network. Most existing approaches usually study networks with a single type of connection between vertices, while in many applications interactions between objects vary, yielding multiplex networks. We propose \model{}, a general, unsupervised edge anomaly detection framework for multiplex dynamic networks. In each relation type, \model{} sees node embeddings at different GNN layers as hierarchical node states and employs a GRU cell to capture temporal properties of the network and update node embeddings over time. We then add an attention mechanism that incorporates information across different types of relations. Our case study on brain networks shows how this approach could be employed as a new tool to understand abnormal brain activity that might reveal a brain disease or disorder. Extensive experiments on nine real-world datasets demonstrate that \model{} achieves state-of-the-art~performance.   
\end{abstract}

\section{Introduction}\label{sec:introduction}

Identifying anomalous activities in networks is a long-standing and vital problem with a wide variety of applications in different domains, e.g., finance, social networks, security, and public health~\cite{anomaly-fraud, anomaly-first, anomaly-survey2, survey_anomaly_detection, anomaly-survey-dynamic}. While several anomaly detection approaches focus on the topological properties of networks~\cite{anomaly_static1, anomaly_static2, anomaly_static3, anomaly_static4, anomaly_static5, anomaly_static6}, detecting anomalies in real-world networks also requires attention to their dynamic nature~\cite{anomaly-survey-dynamic}. Anomalies might appear as malware in computer systems~\cite{unicorn}, social bots and social spammers in social networks~\cite{social_bots}, or financial fraud in financial systems~\cite{anomaly-fraud, anomaly_financial}. Accordingly, anomaly detection in dynamic (evolving) complex systems has recently~attracted~much~attention.

Most prior work effort focuses on detecting anomalies in dynamic networks whose edges are all of the same type~\cite{anomaly_prob1, anomaly_distance1, F-FADE, anomaly_distance3, AddGraph, NetWalk, anomaly_dynamic1, anomaly-survey-dynamic}; these networks are called single-layer, dynamic networks. However, in many complex dynamic systems,
there are many different kinds of interactions between objects. For example, interactions between people can be social or professional, and professional interactions can differ according to topics.
We model graphs with different kinds of edges as Multilayer or Multiplex networks~\cite{main-ML}.
In these networks, the different types of connections are complementary to each other, providing more complex and richer information than simple graphs. Surprisingly, anomaly detection in multiplex networks is relatively less explored and has only recently attracted attention.

Existing approaches to anomaly detection in multiplex networks suffer from three main limitations: \textbf{(1)} Structure and feature inflexibility: existing methods assume pre-defined anomaly patterns or man-made features. Such approaches are application dependent and do not easily generalize to different domains. Moreover, in the real-world networks, anomalies might be more complex in nature, and it is nearly impossible to detect anomalies with high accuracy using pre-defined patterns/roles. \textbf{(2)} Same importance for all type of connections: these methods treat each relation type (i.e., layer) identically, assigning the same importance to each layer. However, real-world multiplex networks can contain noisy/insignificant layers~\cite{FirmCore, ml-core-journal}. Moreover, all vertices might not participate equally in all layers, so which layers are noisy/insignificant can be different for each vertex~\cite{FirmCore, ml-core-journal}. \textbf{(3)} Lack of edge anomaly detection: previous methods for anomaly detection in multiplex networks focus on identifying anomalous nodes, subgraphs, or events. However, in many real-world applications, a connection between two vertices might be an anomaly~\cite{NetWalk, F-FADE, AddGraph, anomaly_prob1}. This anomalous connection might be a suspicious transaction in a financial network, a fake follower in a social network, or an abnormal functional correlation between two regions of the brain.

Existing methods for anomaly detection in single-layer dynamic graphs also exhibit limitations. \textbf{(1)} Structure inflexibility: even in single layer networks, most existing anomaly detection methods for dynamic networks rely on pre-defined patterns or heuristic rules (see \cite{anomaly_distance5, anomaly_distance4}). These heuristic rules are usually content features or long-term temporal factors. However, due the complex nature of real-world anomalies, these factors are not flexible and are restricted in a specific patterns. 
\textbf{(2)} Memory usage: deep learning based methods~\cite{AddGraph, NetWalk}, which are commonly proposed, requiring storing entire snapshots of the network at each time window, consuming large~and~increasing~amounts~of~memory.

To mitigate the above limitations, we introduce \model{} (\textbf{\underline{Ano}}maly Detection in \textbf{\underline{Mul}}tiplex D\textbf{\underline{y}}namic Networks). To take advantage of both temporal properties and complementary information present in multiple relation (edge) types, \model{} extends the idea of \textit{hierarchical node states}~\cite{roland} to multiplex dynamic networks by using an attention mechanism that incorporates information about different relation types. Next, it uses selective negative sampling to learn anomalous edges in an unsupervised manner. To the best of our knowledge, \model{} is the first edge-anomaly detection method for multiplex networks.
Further, when it is possible to model a simple network as a multiplex network, \model{} outperform existing simple network approaches, because the multiplex network provides richer and more complex information than does a simple network~\cite{survey_anomaly_detection, anomaly-survey2, anomaly-survey-dynamic}.

Consider the following two applications for anomaly detection in dynamic multiplex networks:

\head{Applications: Brain Networks}
Monitoring functional systems in the human brain is a fundamental task in neuroscience~\cite{functional_system_brain, functional_system_brain2}. Each node in a brain network represents a region of interest (ROI), which is responsible for a specific function, and edges represent high functional correlation between two ROIs. A temporal brain network is usually derived from functional magnetic resonance imaging (fMRI), which lets us measure the statistical association between the functionality of ROIs over time. Since a (dynamic) brain network generated from an individual can be noisy and incomplete~\cite{Brain_network_fmri, FirmTruss}, prior work used the average of brain networks from many individuals~\cite{anomaly_brain1, brain_dataset}. However, these methods ignore the complex relationships in each individual's brain.
We can capture these missing relationships
by modeling the network as a multiplex (dynamic) network~\cite{ml_brain_first, FirmTruss}, where each layer represents an individual's brain network. We show that edge anomaly detection approaches in a multiplex brain network can be used to detect abnormal connections in the brains of people who live with a brain disease or disorder (see~\autoref{sec:experiments}).

\head{Applications: Fraud Detection in Multiple Blockchain Networks}
Anomaly detection in (dynamic) blockchain transaction networks has recently attracted enormous attention~\cite{blockchain_anomaly1, blockchain_anomaly2, blockchain_anomaly3, blockchain_anomaly4, blockchain_anomaly5, blockchain_anomaly_survey}, due to the emergence of a huge assortment of financial systems' applications~\cite{blockchain_application1, blockchain_application2, blockchain_application3}.  While most existing work focuses on detecting illicit activity in a single blockchain network, recent research shows that cryptocurrency criminals increasingly employ cross-cryptocurrency trades to hide their identity~\cite{crypto_criminals, ofori2021topological}. Accordingly, \citet{blockchain_ml_first} have recently shown that analyzing links across several blockchain networks is critical for identifying emerging criminal activity on the blockchain. An edge anomaly detection approach in multiplex networks can be employed to detect suspicious transactions and identify criminal activities across several blockchain~transaction~networks~more~accurately.

The contributions of this work are: 
\textbf{(1)} We present a novel layer-aware node embedding approach in multiplex dynamic networks, \encoder, which uses an attention mechanism to incorporate both temporal and structural information on different relation types.
\textbf{(2)} We present \model, a general end-to-end unsupervised learning method for anomalous edge detection in multiplex dynamic networks, using a GRU cell to incorporate the outputs of \encoder{} for different snapshots. \textbf{(3)} We demonstrate a new application of edge-anomaly detection in dynamic multiplex networks and present a case study on brain networks of people living with attention deficit hyperactivity disorder (ADHD). Our results show the effectiveness and usefulness of \model{} in identifying abnormal connections of different ROIs of the human brain. This approach could be employed as a new tool to understand abnormal brain activity that might reveal a disease or disorder.
\textbf{(4)} We conduct extensive experiments on nine real-world multiplex and simple networks. Results show the superior performance of \model{} in both single-layer and multiplex networks. 
\section{Related Work}
\label{sec:related-work}
We begin with a brief review of anomaly detection algorithms in dynamic simple networks, then methods for dynamic and multiplex graph learning, and finally anomaly detection in multiplex networks. To situate our research in a broader context, we discuss work on anomaly detection in brain and blockchain networks (~\autoref{app:additional-related-work}). For additional related work, we refer the reader to the extensive survey by \citet{survey_anomaly_detection}.

\head{Anomaly Detection in Dynamic Networks}
There has been much work in anomaly detection for single-layer  dynamic networks. That work falls in five categories: \textbf{(1)} Probabilistic methods~\cite{anomaly_prob1, anomaly_prob2, anomaly_prob3, anomaly_prob4, anomaly_prob5, anomaly_prob6, anomaly_prob7} that identify anomalies based on the pattern deviation from the regular communication patterns. \textbf{(2)} Distance-based methods~\cite{anomaly_distance1, anomaly_distance2, anomaly_distance3, anomaly_distance4, anomaly_distance5} that use certain time-evolving measures of dynamic network structures and use their changes to detect anomalies. \textbf{(3)} Density-based methods~\cite{anomaly_density1, anomaly_density2, anomaly_density3} that view anomalies as subgraphs with high density or as subgraphs with sudden changes in their density. \textbf{(4)} Matrix factorization methods~\cite{matrix_factorization1, matrix_factorization2, matrix_factorization3, matrix_factorization4, matrix_factorization5} that use the low-rank property of network structures and define anomalies as breakers of this property. \textbf{(5)} Learning-based methods~\cite{AddGraph, NetWalk, anomaly_dynamic_transformer, anomaly_dynamic_learning1, anomaly_dynamic_learning2, anomaly_learning3}, that combine the graph embedding method into the anomaly detection approach. These learning-based models must store the entire snapshot, which requires large memory,~limiting~their~scalability. To show the effectiveness of \model{} in even single-layer networks, we compare it with \textsc{NetWalk}\cite{NetWalk}, and \textsc{AddGraph}~\cite{AddGraph} in \autoref{sec:experiments}. All these methods apply to single-layer networks only and do not naturally extend to~multiplex~networks.
We further discuss the novelty of the architecture of our approach in~\autoref{app:additional-related-work}.

\head{Dynamic Graph Neural Networks}
The problem of learning from dynamic networks has been extensively studied in the literature~\cite{time-then-graph, dynamic_gnn1, dynamic_learning1, dynamic_learning2, dynamic_learning3, dynamic_learning4, dynamic_rnn1, yang2022few}. The first group of existing methods use Recurrent Neural Networks (RNN) and then replace its linear layer with graph convolution layer~\cite{dynamic_rnn1, dynamic_rnn2, dynamic_rnn3}.  
The second group uses a GNN as a feature encoder and then deploys a sequence model on top of the GNN to encode temporal properties~\cite{dynamic_gnn1, dynamic_gnn2, dynamic_gnn3}. However, all these models have limitations in both model design and training strategy~\cite{roland}. To address these limitations, \citet{roland} proposed \textsc{ROLAND}, a graph learning framework for dynamic graphs that can re-purpose any static GNN to dynamic graphs. However, this framework cannot be used for graphs with different types of edges (multiplex networks). Our work extends  \textsc{ROLAND} to multiplex networks and introduces an attention mechanism that incorporates the relation-specific hierarchical node states in each snapshot, taking advantage of additional information present in multiplex networks.

\head{Multiplex Graph Learning}
In a multiplex network, also known as multilayer, multi-view, or multi-dimensional networks, all nodes have the same type, but edges (relations) have multiple types~\cite{main-ML}. Several methods have been proposed to learn network embeddings on multiplex networks by integrating information from individual relation type~\cite{multiplex_int1, multiplex_int2, multiplex_int3, hetro1, hetro2, wang2020dynamic}. Other work proposed Graph Convolutional Networks (GCN) methods for multiplex networks~\cite{CS-MLGCN, multiplex_gcn1, multiplex_gcn2, multiplex_gcn1}. Inspired by Deep Graph Infomax~\cite{dgi}, \citet{unsupervisedML-2020} and \citet{multiplex_dgi} proposed unsupervised approaches to learn node embeddings by maximizing the mutual information between local patches and the global representation of the entire graph. \citet{scalable-ml-embedding} proposed a method that uses a latent space to integrate the information across multiple views. Recently, \citet{deep-partial-ml} proposed \textsc{DPMNE} to learn from incomplete multiplex networks. 



\head{Anomaly Detection in Multiplex Networks}
The problem of anomaly detection in multiplex networks has recently attracted attention. \citet{mittal2018anomaly} use eigenvector centrality, page rank centrality, and degree centrality as handcrafted features for nodes to detect anomalies in static multiplex networks. \citet{Bindu2017} proposed a node anomaly detection algorithm in static multiplex networks that uses handcrafted features based on  clique/near-clique and star/near-star structures. \citet{Bansal2020} defined a quality measure, Multi-Normality, which employs the structure and attributes together of each layer to detect attribute coherence in neighborhoods between layers. \citet{Centrality-Based-Anomaly} use 
centrality of all nodes in each layer and apply many-objective optimization with full enumeration based on minimization to obtain Pareto Front. Then, they use Pareto Front as a basis for finding suspected anomaly nodes. \citet{AnoMAN} proposed \textsc{AnomMAN} that uses an auto-encoder module and a GCN-based decoder to detect node anomalies in static multiplex networks. Although this model can learn from the data, it is limited to static networks, and it treats each layer equally in the \textit{Structure Reconstruction} step. 
Finally, \citet{ofori2021topological} developed a new persistence summary and utilized it to detect events in dynamic multiplex blockchain networks.

All of these approaches are designed to detect topological anomalous subgraphs, nodes, or events, and cannot identify anomalous edges. Moreover, as we discussed in \autoref{sec:introduction}, these methods, except \textsc{AnoMAN}~\cite{AnoMAN}, are based on pre-defined patterns/roles or handcrafted features, while in real-world networks anomalies have complex nature. Therefore, these models cannot be generalized to different domains, limiting their application.

\section{Proposed Method: \model}\label{sec:method}

\subsection{Preliminaries}
\label{sec:preliminaries}
We first precisely define multiplex dynamic networks, and then we formalize the problem of edge anomaly detection in multiplex dynamic networks.

\begin{dfn}[Multiplex Dynamic Networks]
    Let $\GG = \{ G_r \}^{\LL}_{r = 1} = (\V, \E, \X)$ denotes a multiplex dynamic network, where $G_r = (\V, \E_r, \X)$ is a graph of the relation type $r \in \LL$ (aka layer), $\V$ is the set of nodes, $\E = \bigcup_{r = 1}^{\LL} \E_r$ is the set of edges, and $\X \in \mathbb{R}^{|\V|\times f}$ is a matrix that encodes node attribute information for nodes in $\V$. Given a relation type $r$, we denote the set of vertices in the neighborhood of $u \in \V$ in relation $r$ as $\mathcal{N}_r(u)$. Each edge $e \in \E$ is associated with an edge type (layer) $r_e \in \LL$ and a timestamp $\tau_e$, and each node $v \in \V$ is associated with a timestamp $\tau_v$.
\end{dfn}

We take a snapshot-based anomaly detection approach for multiplex dynamic networks: a multiplex dynamic graph $\GG = \{\GG^{(t)}\}_{t = 1}^{T}$ can be represented as a sequence of multiplex network snapshots, where each snapshot is a static multiplex graph $\GG^{(t)} = \{ G^{(t)}_r \}_{r = 1}^{\LL} = (\V^{(t)}, \E^{(t)}, \X^{(t)})$ with $\V^{(t)} = \{v \in \V | \tau_v = t \}$ and $\E^{(t)} = \{ e \in \E | \tau_e = t\}$. Our goal is to detect anomalous edges in $\E^{(t)}$. Specifically, for each edge $e = (u, v, r) \in \E^{(t)}$, we produce a layer-dependent anomalous probability $\varphi_r(e)$ in layer $r \in \LL$. 

\begin{figure}
\centering
    \hspace*{-9ex}
    \subfloat[][\centering Architecture of the \textit{Snapshot Encoder}]{{\includegraphics[width=0.71\textwidth]{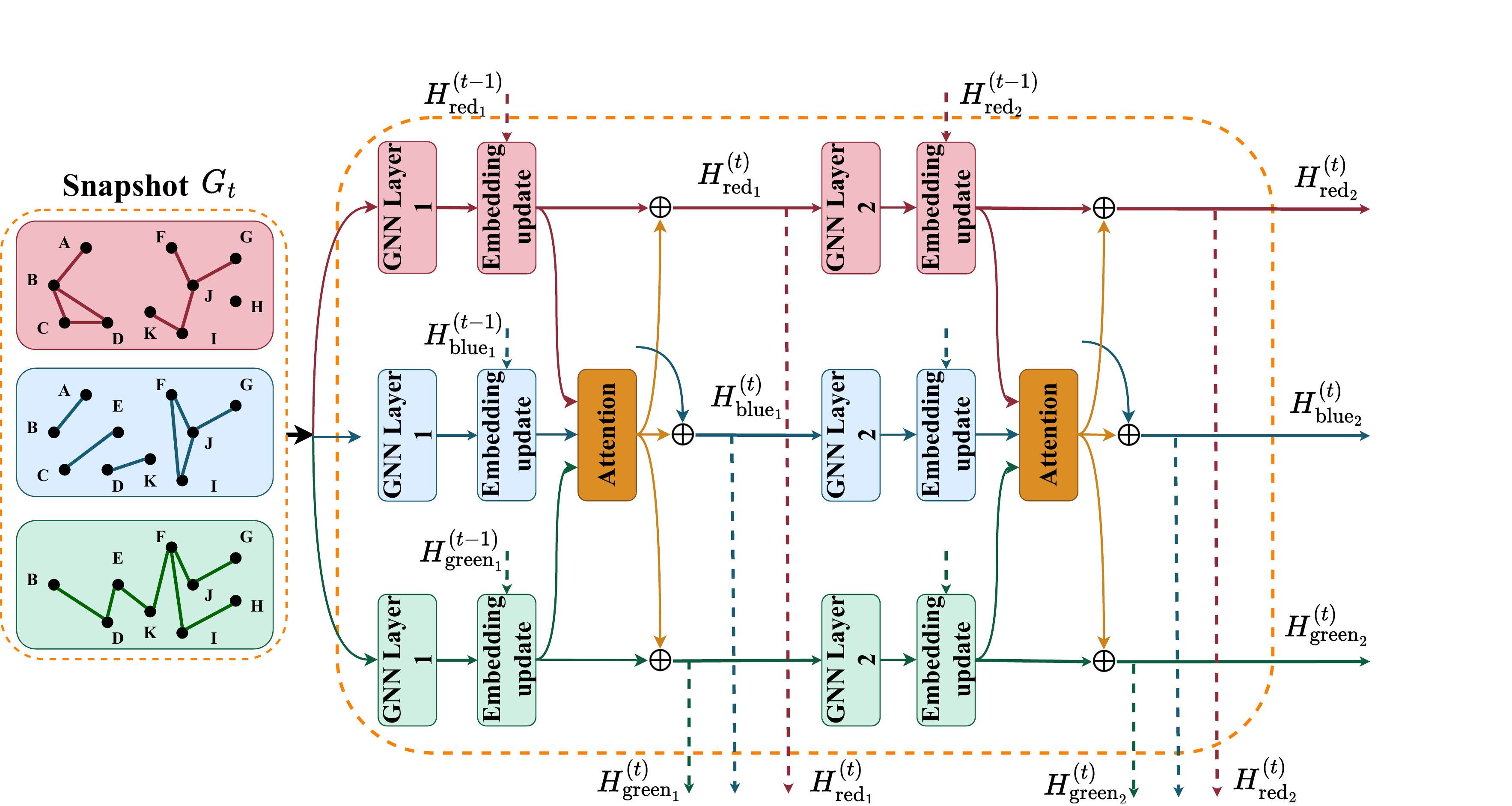} }}
    \hspace*{-8ex}
    \subfloat[][\centering \model{} Framework]{{\includegraphics[width=0.67\textwidth]{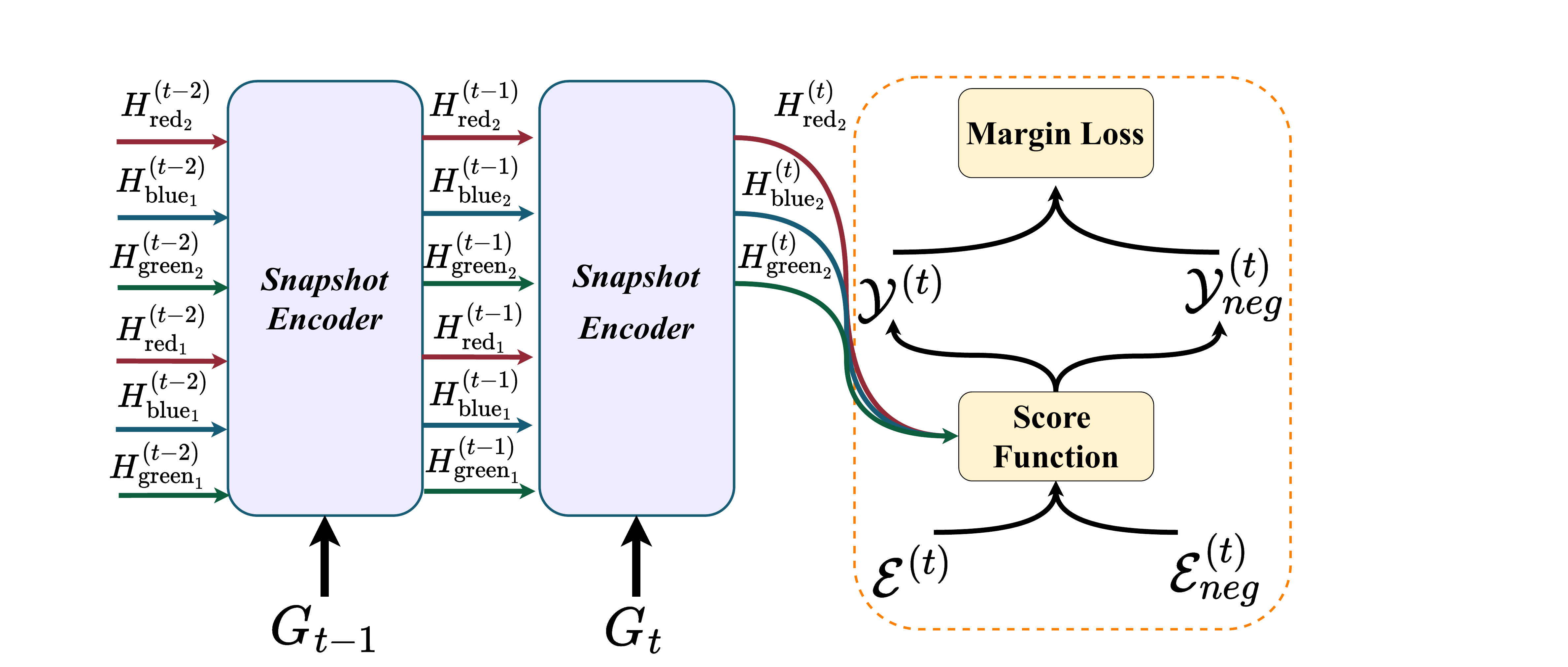} }}
    \caption{Framework and design of \model{} model.}
    \label{fig:model_framework}
    \vspace{-2ex}
\end{figure}

\subsection{\model{} Framework}\label{sec:framework}
We now introduce our framework for edge anomaly detection in dynamic networks with multiple types of interactions. \autoref{fig:model_framework} provides an overview of the framework. To learn the pattern of normal edges, \model{} uses the \encoder{} architecture to encode each snapshot of the network. \encoder{} uses GNNs and incorporates structural and temporal properties of the graph as well as node features in each layer. Next, it uses an attention mechanism to take advantage of complementary information from different layers. Since the graph is dynamically changing, the node embeddings need to change. To this end, \encoder{} uses a GRU cell~\cite{GRU} to update the hierarchical node states over time. Finally, we use the hierarchical node states at each timestamp to calculate the layer-dependent anomalous probabilities of an existing edge and a negative sampled edge and use them as inputs to a margin loss computation. \model{}'s algorithm~appears~in~detail~in~\autoref{app:algorithm}.

\head{GNN Architecture}
A GNN iteratively aggregates messages from the local neighborhood of nodes to learn node embeddings. For a given type of relation $r$, we use the embedding matrix $\tilde{H}_r^{(\ell)} = \{ \tilde{\hh}_{r_{u}}^{(\ell)}\}_{u \in \V}$ to denote the embedding of all vertices in relation type $r$ after applying $\ell$-th GNN layer. Given a relation type $r$, the $\ell$-th layer of the GNN, $\tilde{H}_r^{(\ell)} = \gnn_r(\tilde{H}_r^{(\ell - 1)})$, is defined as:
\begin{equation}
    \begin{aligned}
    &{\mm^{\ell}_{r_{(v \rightarrow u)}}} = W^{(\ell)}_r \textsc{Concat}\left( \tilde{\hh}_{r_{v}}^{(\ell - 1)}, \tilde{\hh}_{r_{u}}^{(\ell - 1)}, \tau_{(v, u, r)} \right), \\
    &\tilde{\hh}_{r_{u}}^{(l)} = \textsc{Agg}^{(\ell)}\left( \left\{  {\mm^{\ell}_{r_{(v \rightarrow u)}}} | v \in \mathcal{N}_r(u)  \right\} \right) + \tilde{\hh}_{r_{u}}^{(\ell - 1)}.
    \end{aligned}
\end{equation}
In our experiments, we follow~\citet{roland}, and use summation as the aggregation function, i.e., $\textsc{Agg}(.) = \textsc{Sum}(.)$. We also use skip-connections~\cite{deepgcn} after aggregation.

\head{Update Modules}
Given the snapshot $\GG^{(t)} = (\V^{(t)}, \E^{(t)}, \X^{(t)})$ at time $t$ and a relation type $r$, we denote the embedding matrix in relation $r$ after the $\ell$-th GNN layer at time $t$ by $\tilde{H}_r^{(t)^{(\ell)}} = \{ \tilde{\hh}_{r_{u}}^{(t)^{(\ell)}}\}_{u \in \V}$. To take advantage of historical data and update the node embeddings at each timestamp, in the \textit{Embedding update} block (see~\autoref{fig:model_framework}), we use a GRU cell~\cite{GRU}. Given a relation type $r$, the output of the $\ell$-th \textit{Embedding update}, $\hat{H}_r^{(t)^{(\ell)}} = \{ \hat{\hh}_{r_{u}}^{(t)^{(\ell)}}\}_{u \in \V}$, is: 
\begin{equation}\label{eq:gru}
    \hat{H}_r^{(t)^{(\ell)}} = \textsc{Gru}_r\left( \tilde{H}_r^{(t)^{(\ell)}}, H_r^{(t - 1)^{(\ell)}} \right),
\end{equation}
where $\tilde{H}_r^{(t)^{(\ell)}}$ is the output of the $\ell$-th GNN layer, and $H_r^{(t - 1)^{(\ell)}}$ is the layer-aware embedding matrix at time $t - 1$.

\head{Attention Mechanism}
The role of our attention mechanism is to incorporate information from different relation types in a weighted manner. As discussed in \autoref{sec:introduction}, the importance of a layer can differ for different nodes, so we cannot calculate a single weight for each layer. Accordingly, we suggest an attention mechanism that learns the importance of layer $r$ for an arbitrary node $u\in \V$. Let $\zeta^{(t)^{(\ell)}}_u$ be the aggregated hidden feature of node $u\in \V$ after the $\ell$-th attention layer at time $t$, we call it a network-level embedding, and $\alpha_{r_u}^{(\ell)}$ indicates the importance of relation type $r$ for vertex $u$, then:  
\begin{equation}
    \zeta^{(t)^{(\ell)}}_u = \sum_{r = 1}^{L} \alpha_{r_u}^{(\ell)} \hat{\hh}_{r_u}^{(t)^{(\ell)}}, 
\end{equation}
where $\hat{\hh}_{r_u}^{(t)^{(\ell)}}$ is the output of $\ell$-th \textit{Embedding update} for node $u$ at time $t$. Following the recent attention-based models~\cite{unsupervisedML-2020, attention_main}, we use the softmax function to define the importance weight of relation type $r$ for node $u$:
\begin{equation}
    \alpha_{r_u}^{(\ell)} = \frac{\exp\left( \sigma\left( {\textbf{s}^{(t)^{(\ell)}}_r}^T \W_{r}^{\text{att}} \:\: \hat{\hh}_{r_u}^{(t)^{(\ell)}} \right) \right)}{\sum_{k = 1}^{L} \exp\left(  \sigma\left( {\textbf{s}^{(t)^{(\ell)}}_k}^T \W_{k}^{\text{att}}  \:\: \hat{\hh}_{k}^{(t)^{(\ell)}} \right)  \right)},
\end{equation}
where $\textbf{s}^{(t)^{(\ell)}}_r$ is a summary of the network in relation type $r$ at time $t$, i.e., $\textbf{s}^{(t)^{(\ell)}}_r = \sum_{u \in \V} \hat{\hh}_{r_u}^{(t)^{(\ell)}}$, and $\W^{\text{att}}_r$ is a trainable weight matrix. In our experiments, we use $\tanh(.)$ as the activation function $\sigma(.)$.

\head{Layer-aware Embedding}
The output of the attention mechanism is a network-level node embedding matrix, which summarizes the properties of nodes over all relation types. Given a relation type $r$, to obtain the layer-aware node embedding of a vertex $u \in \V$, we aggregate the output of the \textit{Embedding update} block, i.e. $\hat{\hh}_{r_u}^{(t)^{(\ell)}}$, and this network-level node embedding, i.e. $\zeta^{(t)^{(\ell)}}_u$. That is:
\begin{equation}\label{eq:layer-aware}
    \hh_{r_u}^{(t)^{(\ell)}} = \textsc{Agg}^{(\ell)} \left( \hat{\hh}_{r_u}^{(t)^{(\ell)}}, \zeta^{(t)^{(\ell)}}_u \right).
\end{equation}
Based on \autoref{eq:layer-aware}, we obtain the layer-aware node embedding matrix, $H_r^{(t)^{(\ell)}} = \{ \hh_{r_{u}}^{(t)^{(\ell)}}\}_{u \in \V}$, for any relation type $r$. Note that we use the layer-aware node embedding matrix at time $t - 1$, $H_r^{(t - 1)^{(\ell)}}$, in \autoref{eq:gru} to update node embeddings after the $\ell$-layer GNN layer.

\head{Anomalous Score Computation}
Now, we get the layer-aware node embedding matrix $H_r^{(t)} = H_r^{(t)^{(L)}}$ at time $t$, for each relation type $r$. Here $L$ is the number of GNN layers. Inspired by~\citet{AddGraph}, for an edge $(u, v) \in \E_r$, we define its anomalous score as follows:
\begin{equation}
    \varphi^{(t)}_r(u, v) = \sigma\left( \eta . \left( || \mathbf{a} \odot \hh_{r_{u}}^{(t)} + \mathbf{b} \odot \hh_{r_{v}}^{(t)} ||^2_2 - \mu  \right) \right),
\end{equation}
where $\sigma(.)$ is an activation function,  $\mathbf{a}$ and $\mathbf{b}$ are trainable vectors, and $\eta$ and $\mu$ are hyperparameters.

\head{Training and Loss Function}
In the training phase, we employ a negative sampling approach in multiplex networks to corrupt edges and generate anomalous connections. Inspired by the negative sampling methods proposed by~\citet{negative_sampling} and \citet{AddGraph}, given a relation type $r$, and a normal edge $(u, v) \in \E_r$, we employ a Bernoulli distribution such that we replace $u$ (resp. $v$) with probability $\frac{deg_r(u)}{deg_r(u) + deg_r(v)}$ (resp. $\frac{deg_r(v)}{deg_r(u) + deg_r(v)}$) in relation type $r$ to generate random negative samples. Since the corrupted edges might be normal, a strict loss function (e.g., cross-entropy) can affect the performance. Accordingly, we employ the margin-based pairwise loss~\cite{loss} in each relation type $r$. Given a relation type $r$, we also employ a $L2$-regularization loss, $\mathscr{L}_r^{reg}$, which is the summation of the $L2$ norm of all trainable parameters, to avoid overfitting. Finally, to aggregate the loss function over all relation types in the multiplex networks,~we~use~the~average~of~loss~functions,~i.e.:

\begin{equation}\label{eq:loss}
    \mathscr{L} =  \min \frac{1}{|\LL|} \left( \sum_{r = 1}^{\LL}   \sum_{(u, v) \in \E_r} \sum_{(u', v') \not \in \E_r} \max \left\{ 0, \gamma + \varphi_r(u, v) - \varphi_r(u', v')  \right\} + \lambda \mathscr{L}_r^{reg} \right).
\end{equation}
Here, $0 \leq \gamma \leq 1$ is the margin between normal and corrupted edges.

\section{Experiments}
\label{sec:experiments}

\head{Experimental Setup and Metrics}
The \encoder{} employs a GNN with 200 hidden dimensions for node states, layers with skip-connection, sum aggregation, and batch-normalization. We tune hyper-parameters by cross-validation on a rolling basis and search the hyper-parameters over \textbf{(i)} the numbers of layers (1 to 5); \textbf{(ii)} learning rate (0.001 to 0.01); and \textbf{(iii)} the margin $\gamma$ (in increments of 0.05 in the range 0.3 to 0.7). The values of the hyper-parameters are reported in \autoref{tab:config} in \autoref{app:config}. We implement \model{} with the GraphGym library~\cite{GraphGym} and use an NVIDIA V100 GPU in the experiments. We use the AUC (the area under the ROC curve) as the metric of comparison. The higher AUC value indicates the high quality of the method.

\begin{table} [tpb!]
\begin{center}
 \caption{Network Statistics.}
    \resizebox{0.78\textwidth}{!} {
\begin{tabular}{l  c c c c c c c c c}
 \toprule
 \multirow{2}{*}{Dataset}   &     \multicolumn{6}{c}{Multiplex Networks} & \multicolumn{3}{c}{Single-layer Networks}\\ 
     \cmidrule(lr){2-7}   \cmidrule(lr){8-10}
   & {RM} &  {DKPol} & {Amazon} &  {Ethereum} & {Ripple} & {DBLP} & {Bitcoin} & {Amazon-S} & {DBLP-S} \\
 \midrule
 \midrule
    $|\V|$  &    91    & 490      &   17.5K   &    221K  &  54K  & 513K  & 3.7K & 8.6K & 23K \\
    $|\E|$     &   14K   &  20K    &  282K   &   473K & 837K & 1M & 24.1K & 90K & 95.2K \\
    $|\LL|$     &   10    &   3    &  2     &   6  &   5  & 10 & 1 & 1 & 1\\
 \bottomrule
\end{tabular}
}
\vspace{-2ex}
 \label{tab:datastat}
 \end{center}
\end{table}

\head{Datasets}
We use nine real-world public datasets~\cite{RM, FirmTruss, ofori2021topological, DKPol, bitcoin_alpha, amazon_dataset_main} whose domains cover social, co-authorship, blockchain, and co-purchasing networks.
We summarize their statistics in \autoref{tab:datastat} and provide detailed descriptions in~\autoref{app:datasets-details}. Since the ground truth for anomaly detection is difficult to obtain~\cite{anomaly-survey2}, we extend the methodology in existing studies~\cite{NetWalk, AddGraph, anomaly-survey2} to multiplex networks and propose a new approach to inject anomalous edges into our datasets (more details in \autoref{app:datasets-details}). Note that the DKPol and RM datasets are static multiplex networks, and we use them to show the effectiveness of \model{} in capturing content and structural features. 

\head{Baselines}
Since there is no prior work on edge anomaly detection in multiplex networks, we first compare \model{} with single-layer edge anomaly detection methods: GOutlier~\cite{anomaly_prob2} builds a generative model for edges in a node cluster. CM-Sketch~\cite{anomaly_distance4} uses a Count-Min sketch for approximating global and local structural properties. NetWalk~\cite{NetWalk} uses a random walk to learn a unified embedding for each node and then dynamically clusters the nodes' embeddings. AddGraph~\cite{AddGraph} is an end-to-end approach that uses an extended GCN in temporal networks.
Finally, we  compare with two multiplex network embedding baselines, ML-GCN~\cite{CS-MLGCN} and MNE~\cite{scalable-ml-embedding}. We apply $K$-means clustering on their obtained node embeddings for anomaly detection~\cite{NetWalk}.

\vspace*{-3ex}
\begin{center}
\begin{table} [tpb!]
    \caption{Performance comparison in multiplex networks (AUC).}
    \begin{center}
\resizebox{1\textwidth}{!} {
\begin{tabular}{l   c c  c c  c c  c c  c c c c}
 \toprule
  Methods & \multicolumn{2}{c}{RM} & \multicolumn{2}{c}{DKPol} & \multicolumn{2}{c}{Amazon} & \multicolumn{2}{c}{DBLP} & \multicolumn{2}{c}{Ethereum} & \multicolumn{2}{c}{Ripple} \\
  \cmidrule(lr){2-3} \cmidrule(lr){4-5} \cmidrule(lr){6-7} \cmidrule(lr){8-9} \cmidrule(lr){10-11} \cmidrule(lr){12-13}
  Anomaly \% & 1$\%$ & 5 $\%$ & 1$\%$ & 5 $\%$ & 1$\%$ & 5 $\%$ & 1$\%$ & 5 $\%$ & 1$\%$ & 5 $\%$ & 1$\%$ & 5 $\%$  \\
 \midrule
 \multicolumn{13}{c}{Single-layer Methods}\\
 \midrule
    \textsc{GOutlier}   & 0.7138 & 0.6982 & 0.6844 & 0.6597 & 0.6973          & 0.6672 & 0.7059 & 0.6901 & 0.7017 & 0.6799 & 0.7036 & 0.6851   \\
    \textsc{CM-Sketch}  & 0.7127 & 0.7012 & 0.7058 & 0.6930 & 0.6881          & 0.6719 & 0.7186 & 0.6915 & 0.7408 & 0.7277 & 0.7360 & 0.7194   \\
    \textsc{NetWalk}    & 0.7739 & 0.7641 & 0.7706 & 0.7581 & 0.7228          & 0.7122 & 0.7742 & 0.7523 & 0.7956 & 0.7885 & 0.7904 & 0.7823   \\
    \textsc{AddGraph}   & 0.8005 & 0.8093 & 0.8149 & 0.8087 & 0.7796          & 0.7735 & 0.8024 & 0.7995 & 0.8133 & 0.8090 & 0.8205 & 0.8217   \\
    \midrule
 \multicolumn{13}{c}{Multiplex Methods}\\
    \midrule
    \textsc{MNE}     & 0.7994 & 0.7955 & 0.8050 & 0.7913 & 0.7108 & 0.7017 & 0.7532 & 0.7499 & 0.7541 & 0.7495 & 0.7813 & 0.7754   \\
    \textsc{ML-GCN}           & 0.7921 & 0.7886 & 0.7915 & 0.7907 & 0.7344 &  0.7263   & 0.7519 & 0.7439 & 0.7940 & 0.7918 & 0.8115 & 0.8072   \\
    \model{}          & \textbf{0.8783} & \textbf{0.8729} & \textbf{0.8694} & \textbf{0.8610} & \textbf{0.8289} & \textbf{0.8195} & \textbf{0.8825} & \textbf{0.8754} & \textbf{0.8906} & \textbf{0.8852} & \textbf{0.8938} & \textbf{0.8871}   \\
    \midrule
    \midrule
    \multirow{2}{*}{Improvement}      &  \multirow{2}{*}{9.71\%} &  \multirow{2}{*}{7.85\%} &  \multirow{2}{*}{6.68\%} &  \multirow{2}{*}{6.46\%} &  \multirow{2}{*}{6.32\%} &  \multirow{2}{*}{5.92\%} &  \multirow{2}{*}{9.98\%} &  \multirow{2}{*}{9.49\%} &  \multirow{2}{*}{9.50\%} &  \multirow{2}{*}{9.41\%} &  \multirow{2}{*}{8.93\%} &  \multirow{2}{*}{7.96\%}   \\
    & \\
  \toprule
\end{tabular}
}
 \label{tab:ML-results}
 \vspace{-3mm}
 \end{center}
\end{table}
\end{center}
\head{Results on Multiplex Networks}
We compare \model{} with the baseline methods on both dynamic and static multiplex networks with different percentages of anomalous edges (i.e., $1\%$, $5\%$). \autoref{tab:ML-results} reports the AUC for both the baselines and \model{}. Our method outperforms \emph{all} baselines in all datasets and improves the best baseline results by $8.18\%$ on average. There are three reasons for \model{}'s superior performance: \textbf{(1)} it outperforms competitors for static datasets because it can learn structural anomaly patterns in the network, rather than depending on pre-defined patterns/roles. \textbf{(2)} \model{} outperforms single-layer methods due to its attention mechanism that incorporates complementary information from different relation types. \textbf{(3)} \model{} outperforms multiplex methods as it is an end-to-end method and is optimized for anomaly detection. It is also a dynamic method and can take advantage of the temporal properties of the network.

\begin{table} [tpb!]
    \parbox{.60\linewidth}{        
        \centering
            \caption{Performance comparison in simple networks~(AUC).}
            \resizebox{0.58\textwidth}{!} {
                    \begin{tabular}{l  c c c c c  c}
                         \toprule
                          Methods & \multicolumn{2}{c}{Bitcoin} & \multicolumn{2}{c}{Amazon-S} & \multicolumn{2}{c}{DBLP-S} \\
                          \cmidrule(lr){2-3} \cmidrule(lr){4-5} \cmidrule(lr){6-7}
                          Anomaly \% & 1$\%$ & 5 $\%$ & 1$\%$ & 5 $\%$ & 1$\%$ & 5 $\%$  \\
                         \midrule 
                         \midrule
                            \textsc{GOutlier}   & 0.7143 & 0.7091 & 0.6923 & 0.6614 & 0.7108 & 0.6995\\
                            \textsc{CM-Sketch}  & 0.7146 & 0.7015 & 0.7049 & 0.6621 & 0.7084 & 0.6877 \\
                            \textsc{NetWalk}    & 0.8375 & 0.8367 & 0.7483 & 0.7302 & 0.7779 & 0.7590  \\
                            \textsc{AddGraph}   & 0.8534 & 0.8416 & 0.7872 & 0.7828 & 0.7911 & 0.7932 \\
                            \model{} & \textbf{0.8707} & \textbf{0.8661} & \textbf{0.8014}& \textbf{0.7943} &  \textbf{0.8129} &  \textbf{0.8236} \\
                          \toprule
                    \end{tabular}
            }
            \label{tab:SL-results}
    }
    \parbox{.405\linewidth}{        
        \centering
            \vspace{-6.5mm}
            \caption{Ablation study (AUC).}
            \resizebox{0.405\textwidth}{!} {
                    \begin{tabular}{l  c c c}
                         \toprule
                          Methods & Amazon & DBLP & Ethereum  \\
                         \midrule 
                         \midrule
                         \model{} & \textbf{0.8289} & \textbf{0.8825} & \textbf{0.8906} \\
                            w/ MLP            & 0.8023 & 0.8606 & 0.8718 \\
                            w/o Attention     & 0.7831 & 0.8219 & 0.8275\\
                            w/ Summation    & 0.8007 & 0.8571 & 0.8688 \\
                          \toprule
                    \end{tabular}
            }
            \label{tab:ablation_study}
    }
    \vspace{-2ex}
\end{table}

\head{Results on Single-layer Dynamic Networks}
We also compare \model{} with single-layer method baselines on single-layer dynamic datasets. \autoref{tab:SL-results} summarizes the results. Once again, we see that \model{} outperforms all the baselines, even in single-layer graphs, by $2.46\%$ on average. This is mainly due to \encoder{}'s architecture, which enables our method to incorporate the outputs of GNN layers after each layer and recurrently update them over time by a GRU cell.

\head{Ablation Study}
Next, we conduct experiments to show that the \model{} architecture design is effective in boosting performance. We examine the effect of GRU cells by replacing them with a 2-layer MLP. We also investigate the effect of the attention mechanism by \textbf{(1)} removing the attention, learning node embeddings in each relation-type separately and \textbf{(2)} aggregating the information of different relation types by summation (without weights). \autoref{tab:ablation_study} summarizes the results. We found that both our attention mechanism and the GRU cells are important for \model{}, producing significant performance boosts.

Additional experimental results on the effect of the training ratio can be found in \ref{app:additional_experiment}.

\begin{figure}[t]
\hspace*{-7ex}
\parbox{.83\linewidth}{
    \centering
    \includegraphics[width=0.95\linewidth]{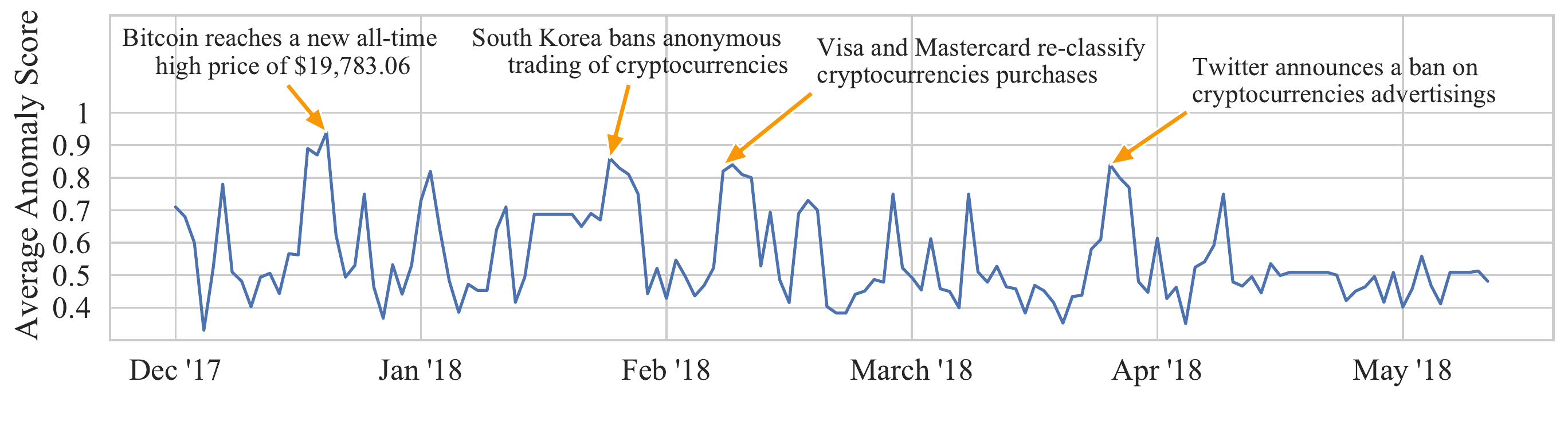}
    \vspace{2ex}
    \caption{Event detection in Ethereum network.}
    \label{fig:event_detection}
}~
\hspace*{-15ex}~\vspace{-1ex}~
\parbox{.50\linewidth}{
\centering
    \hspace{3ex}
    \vspace{-1mm}
    \includegraphics[width=0.4\linewidth]{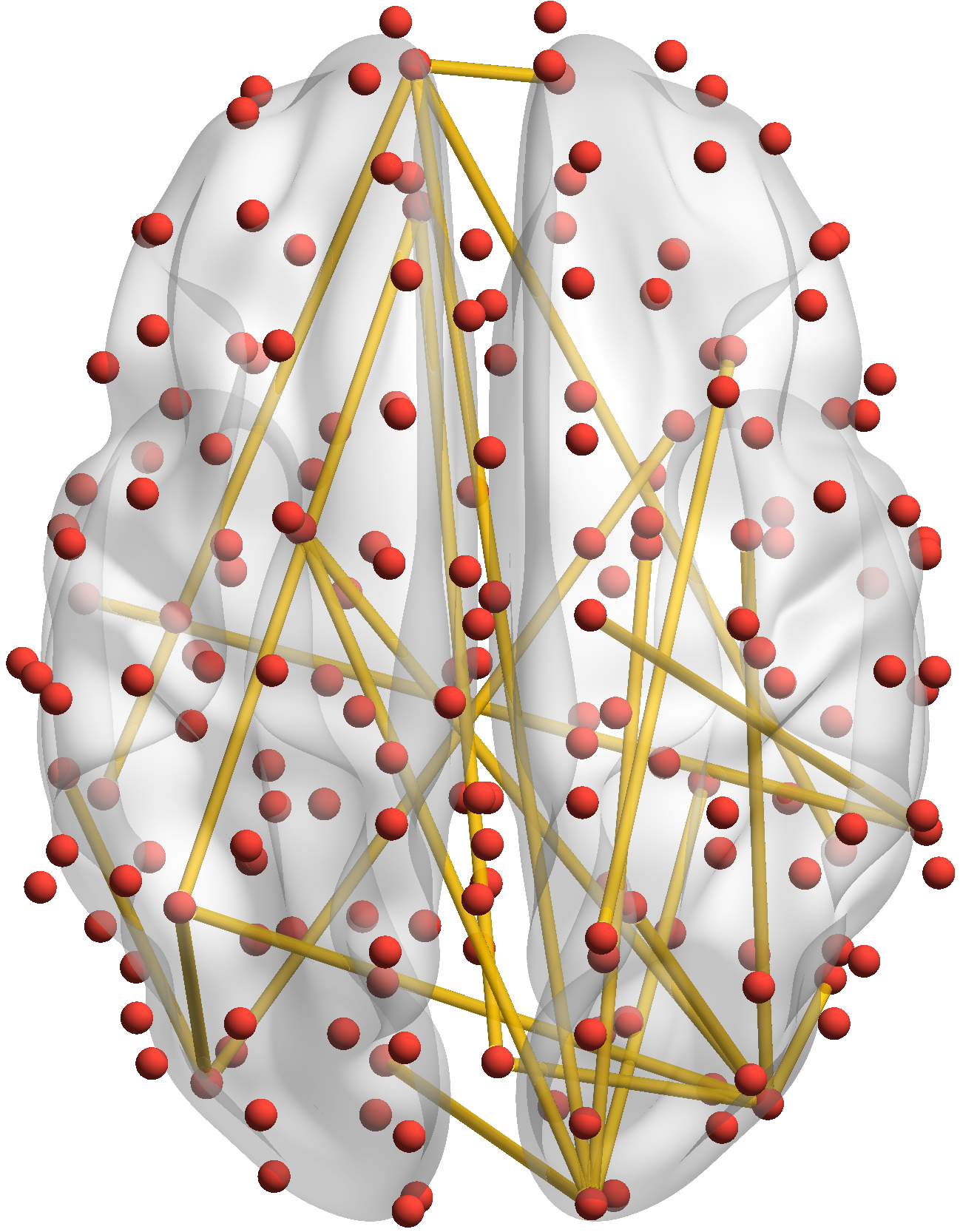}
    \caption{Anomalous edges in BN.}
    \label{fig:brain}
}
\vspace{-2ex}
\end{figure}

\head{Effectiveness in Detecting Events}
Next, we evaluate how well \model{} detects events in the Ethereum transaction network. In each timestamp, we calculate the anomaly score for all the edges in the snapshot. We then compute the average of the top-15 edge anomaly scores and report them in \autoref{fig:event_detection}. We find that the top-4 local optimums all coincide with major events annotated in the figure. 

\head{Case Study of Brain Networks}
Behavioral disturbances in ADHD are thought to be caused by the dysfunction of spatially distributed, interconnected neural systems~\cite{SVM_ADHD}. We used \model{} to detect anomalous connections in the brain network (BN) of people with ADHD. Anomalous functional correlations between BN of people  with ADHD compared to those without can help us understand which brain regions are involved in ADHD. Our dataset~\cite{brain_main_dataset} is derived from the functional magnetic resonance imaging (fMRI) of 40 individuals (20 individuals in the condition group, labeled  ADHD, and 20 individuals in the control group, labeled  TD) with the same methodology used by~\citet{Brain_network_fmri}. Here, each layer (relation type) is the BN of an individual person, where nodes are brain regions, and each edge measures the statistical association between the functionality of its endpoints. We present two results: \textbf{(1)} 74\% of all detected anomalies are edges in the BNs of people in ADHD group. \textbf{(2)} 69\% of all found anomalies in the ADHD group correspond to edges in the frontal and occipital cortex of the brain. \autoref{fig:brain} visualizes the anomalous edges in the brain network of an individual in the ADHD group. These findings show an unexpected functional correlation of occipital and frontal lobes regions with other parts, which are consistent with previous studies on ADHD~\cite{brain_result, anomaly_brain1, brain_result}. More results and visualizations are reported in~\autoref{app:additional_experiment}.

\vspace{-1ex}
\section{Conclusion, Limitations, and Future Work}\label{sec:conclusion}
We present \model{}, an end-to-end unsupervised framework for detecting edge anomalies in dynamic multiplex networks. \model{} is based on a new architecture that employs GNN and GRU cells to take advantage of both temporal and structural properties and adds an attention mechanism that effectively incorporates information across different types of connections. Finally, it uses a negative sampling approach in training to overcome the lack of ground truth data. Extensive experiments show the power of \model{} to effectively identify temporal and structural anomalies in both single-layer and multiplex networks. Our case studies on brain networks and blockchain transaction networks show the usefulness of \model{} in a wide array of applications and domains. The success raises many interesting directions for future studies: \textbf{(1)} \model{} shows the potential to detect anomalous connections in human brains, which can help predict brain diseases or disorders. One future direction is to design an end-to-end model based on the \model{} architecture and directly optimize it to predict the presence of disease. \textbf{(2)} \model{} assigns the same importance to all snapshots, while earlier information might have less impact and importance than more recent information. 

\acksection{}
This research was enabled in part by support provided by WestGrid (\href{https://www.westgrid.ca}{https://www.westgrid.ca}) and The Digital Research Alliance (\href{https://alliancecan.ca/en}{https://alliancecan.ca/en}).
We acknowledge the support of the Natural Sciences and Engineering Research Council of Canada (NSERC).

\bibliographystyle{unsrtnat}
\setcitestyle{numbers}

\bibliography{main}

\newpage
\appendix

\section{Additional Related Work}\label{app:additional-related-work}

\head{Novelty of the \textit{Snapshot Encoder} Architecture}
In the literature, several studies combine GNNs with recurrent models like GRU cells or Transformers. Here, we want to emphasize that the architecture of \textit{Snapshot Encoder} is quite different from them in different aspects. As we discussed in \autoref{sec:related-work}, we can categorize previous works on the combination of GRU and GNNs into two groups. The first group replaces RNN's linear layer with graph convolution layer~\cite{dynamic_rnn1, dynamic_rnn2, dynamic_rnn3}.  
The second group uses a GNN as a feature encoder and then deploys a sequence model on top of the GNN to encode temporal properties~\cite{dynamic_gnn1, dynamic_gnn2, dynamic_gnn3}, which ignores the evolution of lower-level node embeddings. Also, these models have limitations in training strategy~\cite{roland}. ROLAND~\cite{roland} has addressed these limitations, but it is limited to single-layer graphs. Moreover, natural attempts to use multiplex graph neural networks~\cite{CS-MLGCN, multiplex_int1, multiplex_int2, multiplex_int3, multiplex_gcn1, multiplex_gcn2} in the ROLAND framework (e.g. replace GNN block with multiplex GNN or GCN) lead to ignoring historical data in other relation types (layers). For example, assume that we want to use our attention mechanism (see~\autoref{sec:method}) in the ROLAND framework. Then, we need to use the attention mechanism in the GNN block, which means we incorporate the information about different relation types before embedding the \textit{Embedding update} module. Accordingly, for each timestamp, we do not incorporate historical data on other relation types, which could result in undesirable performance.

\head{Feature Learning and Anomaly Detection in Brain Networks}
In recent years, several studies focused on analyzing brain networks to understand and distinguish healthy and diseased human brains~\cite{nongnn_brain_network1, nongnn_brain_network2, nongnn_brain_network3}. Recently, due to the success of GNNs in analyzing graph-structured data, deep models have been proposed to predict brain diseases by learning the graph structures of brain networks\cite{classification_brain_network, classification_brain_network2, classification_brain_network3, classification_brain_network4, classification_brain_network5}. In addition to predicting disease in brain networks, understanding the cause of the disease is important. To this end, several anomaly detection methods have been proposed to find anomalous connections, regions, or subgraphs in the brain, which can cause a disease~\cite{anomaly_brain, anomaly_brain1, anomaly_brain2}. However, all these anomaly detection models apply to single-layer networks only and do not naturally extend to multiplex networks; while a brain network generated from an individual can be noisy and incomplete \cite{Brain_network_fmri, FirmTruss}. To the best of our knowledge, \model{} is the first method for detecting  anomalous connections in multiplex brain networks.

\head{Anomaly Detection in Blockchain Networks}
Anomaly detection in blockchain transaction networks has recently attracted enormous attention~\cite{blockchain_anomaly1, blockchain_anomaly2, blockchain_anomaly3, blockchain_anomaly4, blockchain_anomaly5, blockchain_anomaly_survey}, due to the emergence of a huge assortment of financial systems' applications~\cite{blockchain_application1, blockchain_application2, blockchain_application3}.  However, most existing work focuses on detecting illicit activity in a single blockchain network, while recent research shows that cryptocurrency criminals increasingly employ cross-cryptocurrency trades to hide their identity~\cite{crypto_criminals, ofori2021topological}. Accordingly, \citet{blockchain_ml_first} has recently shown that analyzing links across several blockchain networks is critical for identifying emerging criminal activity on the blockchain. To this end, \citet{ofori2021topological} developed a new persistence summary and utilized it to detect events in dynamic multiplex blockchain networks. For additional related work, we refer the reader to the extensive survey by \citet{blockchain_anomaly_survey}. To the best of our knowledge, \model{} is the first method for detecting anomalous transactions in multiple blockchain networks.

\section{\model{} Algorithm}\label{app:algorithm}
\model{}'s algorithm~appears~in~detail~in Algorithm~\ref{alg:model}.
\begin{algorithm}
    \small
    \caption{\model{} Algorithm}
    \label{alg:model}
    \begin{algorithmic}[1]
        \Require{A multiplex dynamic network $\GG = \{ \GG^{(t)} \}_{t = 1}^{T}$}
        \Ensure{Node embeddings for all relation types $\{H_r^{(T)}\}_{r = 1}^{\LL}$}
        \State Initialize $\left\{ \left\{ H_r^{{(0)}^{(\ell)}} \right\}_{\ell = 1}^{L} \right\}_{r = 1}^\LL$;
        \While{\text{not converged}}
                \For{$t = 1, \dots, T$}
                     \For{$r = 1, \dots, \LL$}
                        \State Let $\mathscr{L}_r^{(t)} = 0$;
                            \For{$\ell = 1, \dots, L$}
                                \State $\tilde{H}_{r}^{{(t)}^{(\ell)}} = \textsc{GNN}_r\left( H_{r}^{(\ell - 1)} \right)$;
                                \State $\hat{H}_r^{(t)^{(\ell)}} = \textsc{Gru}_r\left( \tilde{H}_r^{(t)^{(\ell)}}, H_r^{(t - 1)^{(\ell)}} \right)$;
                                \State $Z_r^{{(t)}^{(\ell)}} = \{\zeta^{(t)^{(\ell)}}_u\}_{u \in \V} = \left\{\sum_{r = 1}^{L} \alpha_{r_u}^{(\ell)} \hat{\hh}_{r_u}^{(t)^{(\ell)}}\right\}_{u \in \V}$;
                                \State $H_{r}^{(t)^{(\ell)}} = \textsc{Agg}^{(\ell)} \left( \hat{H}_{r}^{(t)^{(\ell)}}, Z_r^{{(t)}^{(\ell)}} \right)$
                            \EndFor
                            \For{$(u, v) \in \E^{(t)}_r$}
                                    \State Sample $(u', v')$ for $\varphi^{(t)}_r(u, v)$;
                                    \State $ \mathscr{L}_r^{(t)} = \mathscr{L}_r^{(t)} + \max\left\{ 0, \gamma + \varphi^{(t)}_r(u, v) - \varphi^{(t)}_r(u', v') \right\}$;
                                \EndFor
                            \State $\mathscr{L}_r^{(t)} = \mathscr{L}_r^{(t)} + \lambda \mathscr{L}_r^{reg}$;
                     \EndFor
                    \State $\mathscr{L}^{(t)} = \frac{1}{\LL} \left( \sum_{r = 1}^{\LL} \mathscr{L}_r^{(t)} \right)$;
                    \State Minimize $\mathscr{L}^{(t)}$; 
                \EndFor
        \EndWhile
        \Return $\{H_r^{(T)}\}_{r = 1}^{\LL}$
    \end{algorithmic}
\end{algorithm}

\section{Experimental Configuration}\label{app:config}
\begin{table} [tpb!]
\begin{center}
 \caption{The value of hyper-parameters of \model}
    \resizebox{0.90\textwidth}{!} {
\begin{tabular}{l  c c c c c c c c c}
 \toprule
   \multirow{2}{*}{Dataset}   &     \multicolumn{6}{c}{Multiplex Networks} & \multicolumn{3}{c}{Single-layer Networks}\\ 
     \cmidrule(lr){2-7}   \cmidrule(lr){8-10} 
   & {DKPol} &  {RM} & {Amazon} &  {DBLP} & {Ethereum} & {Ripple} & {Bitcoin} & {Amazon-S} & {DBLP-S} \\
 \midrule
 \midrule
    $\eta$    &  1   & 1   & 1   & 3  & 1   & 1   & 1   & 1   & 3\\
    $\mu$     &  0.3 & 0.25 & 0.5 & 0.5& 0.3 & 0.3 & 0.3 & 0.5 & 0.5\\
    $\lambda$     &  $5\times 10^{-7}$ & $5\times 10^{-7}$ & $5\times 10^{-7}$ & $5\times 10^{-7}$ & $5\times 10^{-7}$ & $5\times 10^{-7}$ & $5\times 10^{-7}$ & $5\times 10^{-7}$ & $5\times 10^{-7}$\\
 \bottomrule
\end{tabular}
}
 \label{tab:config}
 \end{center}
\end{table}

In the architecture of \encoder{} we use 200 hidden dimensions for node states, GNN layers with skip-connection, sum aggregation, and batch-normalization. We tune hyper-parameters by cross-validation on a rolling basis, and search the hyper-parameters over \textbf{(i)} numbers of layers (1 layer to 5 layers); \textbf{(ii)} learning rate (0.001 to 0.01); and \textbf{(iii)} the margin $\gamma$ (0.3 to 0.7). The values of other hyper-parameters are reported in \autoref{tab:config}.

\section{Datasets}\label{app:datasets-details}
We use nine real-world public datasets~\cite{RM, FirmTruss, ofori2021topological, DKPol, bitcoin_alpha, amazon_dataset_main} whose domains cover social, co-authorship, blockchain, and co-purchasing networks. We summarize their statistics in \autoref{tab:datastat}. 

\head{Social Networks} RM~\cite{RM} has 10 networks, each with 91 nodes. Nodes represent phones and one edge exists if two phones detect each other under a mobile network. Each network describes connections between phones in a month. DKPol~\cite{DKPol} is collected during the month leading to the 2015 Danish parliamentary election on Twitter. Nodes are Twitter accounts of Danish politicians, and relations are "Retweet", "Reply", and "Topical Interaction"~\cite{DKPol}.

\head{Co-purchasing Network} Amazon~\cite{amazon_dataset_main} is a co-purchasing network, where each node is an item and the type of connections are "Also-view" and "Also-bought". We focused on items with four categories, i.e., Beauty, Automotive, Patio Lawn and Garden, and Baby. 

\head{Collaboration Network} DBLP is a co-authorship network until 2014 from \cite{dblp}. In this dataset, each node is a researcher, an edge shows collaboration and each type of connection is a topic of research. For each collaboration, we consider the bag of words drawn from the titles of the paper and apply LDA topic modeling~\cite{LDA} to automatically identify 240 topics. We then cluster their non-zero elements into ten known research topics.

\head{Blockchain Networks} Ethereum~\cite{ofori2021topological} is a blockchain transaction network over 576 days, where layers are different tokens, nodes are addresses of investors, and edge denotes the transferred token value. Since the same address may trade multiple tokens, the address appears in networks of all the tokens it has traded. Ripple~\cite{ofori2021topological} is derived from the Ripple Credit Network and covers a timeline of Oct-2016 to Mar-2020. Similar to the Ethereum dataset, nodes are investors and edges represent transactions. Here layers (relation types) correspond to the five most issued fiat currencies on the Ripple network: JPY, USD, EUR, CCK, and CNY.

\head{Brain Network} We use ADHD-Brain dataset in our case study. This dataset~\cite{brain_main_dataset} is derived from the functional magnetic resonance imaging (fMRI) of 40 individuals (20 individuals in the condition group, labeled  ADHD, and 20 individuals in the control group, labeled  TD) with the same methodology used by~\citet{Brain_network_fmri}. Here, each layer (relation type) is the brain network of an individual person, where nodes are brain regions, and each edge measures the statistical association between the functionality of its endpoints.

\head{Single-layer Networks} DBLP-S is a subgraph of the multiplex DBLP network, where we collect a subset of researchers who work on data mining and related areas. Amozon-S, is a subgraph of the multiplex Amazon network, where we focus on the "Also-view" relation type. Bitcoin dataset contains who-trusts-whom network of people who trade on the Alpha platforms~\cite{bitcoin_alpha}.

Note that all of the datasets are anonymized and does not contain any personally identifiable information or offensive content.

\subsection{Inject Anomalous Edges in Multiplex Networks}
\label{app:inject_anomaly}
Since the ground truth for anomaly detection is difficult to obtain~\cite{anomaly-survey2}, we follow the existing studies\cite{NetWalk, AddGraph, anomaly-survey2} and inject anomalous edges into our datasets. However, existing methods only inject anomalous edges in a single-layer network and cannot add complex anomalies in multiplex networks. Accordingly, here we divide injected edges into two groups (50\% each), \textbf{(1)} layer-independent anomalies, and \textbf{(2)} layer-dependent anomalies. For the first group, we use existing methods~\cite{anomaly-survey2}. In multiplex networks, the rich information about node connections leads to repetitions, meaning edges between the same pair of nodes repeatedly appear in multiple layers. Repeated connections are more likely to be a strong tie and even its nodes belong to the same community \cite{Redundancy-ML-Community, FirmTruss}. Accordingly, in the second type of injected anomalies, we inject random connections that do not appear in any relation type. That is, we first choose a random edge $(u, v)$ such that $(u, v) \not\in \E_k$ for all $k \in \LL$, and then we inject this edge to a random relation type $r \in \LL$. Since this connection does not appear in any relation type, it is more likely to be an anomaly. This type of anomaly helps us to understand whether \model{} can take advantage of complementary information of different relation types.

\section{Additional Experimental Results}\label{app:additional_experiment}

\subsection{Parameter Sensitivity}
\begin{figure}[t]
    \centering
    \includegraphics[width=0.7\linewidth]{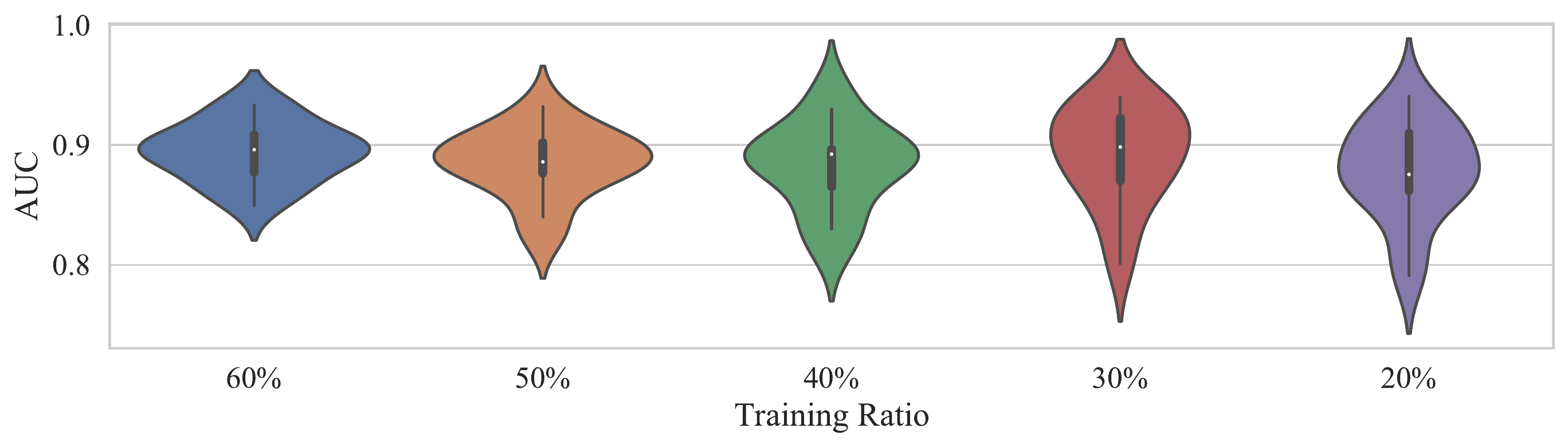}
    \vspace{-1mm}
    \caption{Stability over different training ratio on Ripple.}
    \label{fig:training_ratio}
\end{figure}
We evaluate the effect of the training ratio of the dataset: we change the training ratio from 60\% to 20\% and report the results on the Ripple dataset in \autoref{fig:training_ratio}. Decreasing the training ratio tends to increase both the average and maximum AUC, except for the 20\% case,
and decreases the minimum AUC for all training ratios.
These results show that performance stays relatively stable, which demonstrates that our framework is robust in the presence of a small amount of training data.

\begin{figure}
\centering
    \subfloat[][\centering Left View]{{\includegraphics[width=0.35\textwidth]{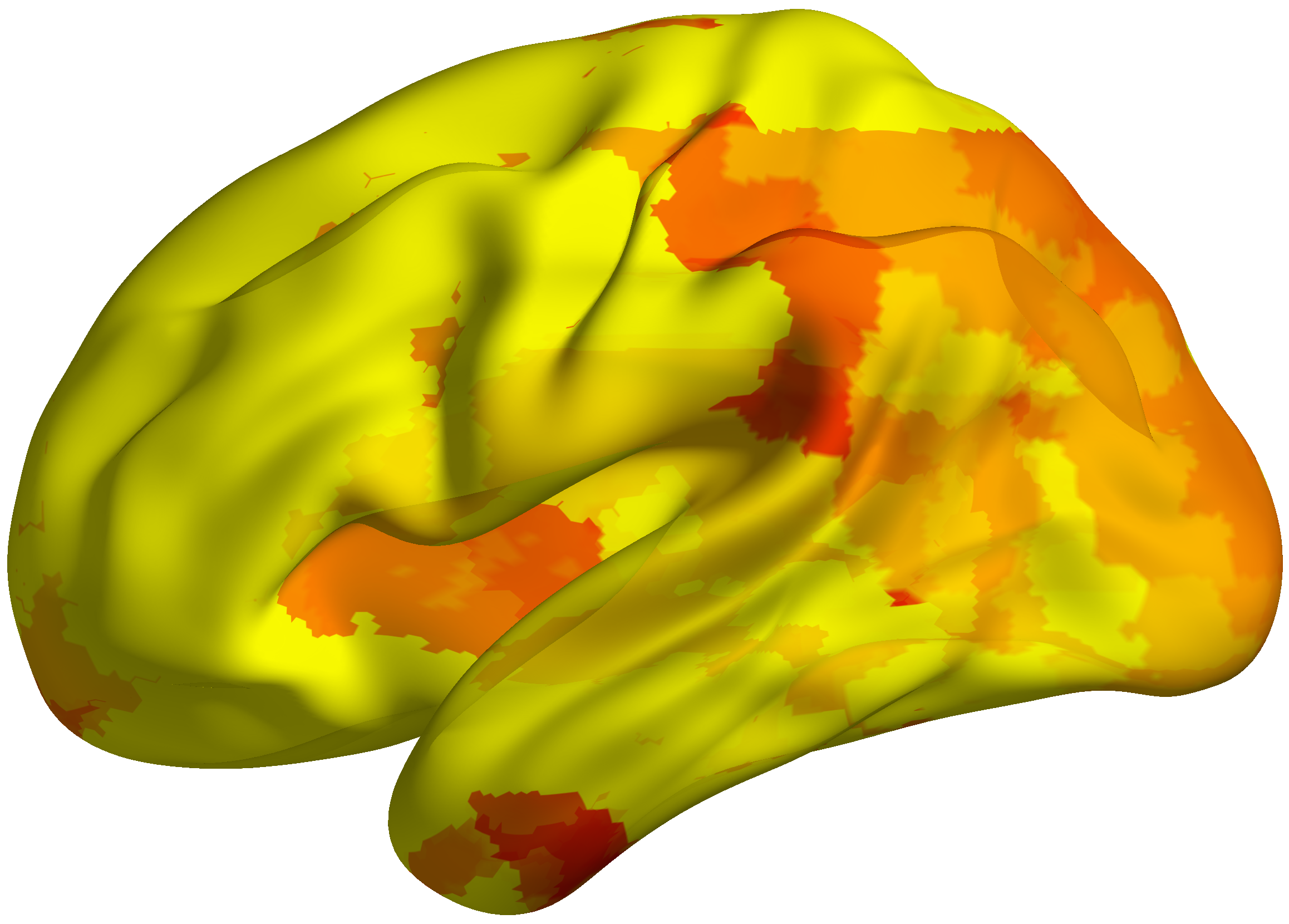} }}
    \hspace{1ex}~
    \subfloat[][\centering Right View]{{\includegraphics[width=0.35\textwidth]{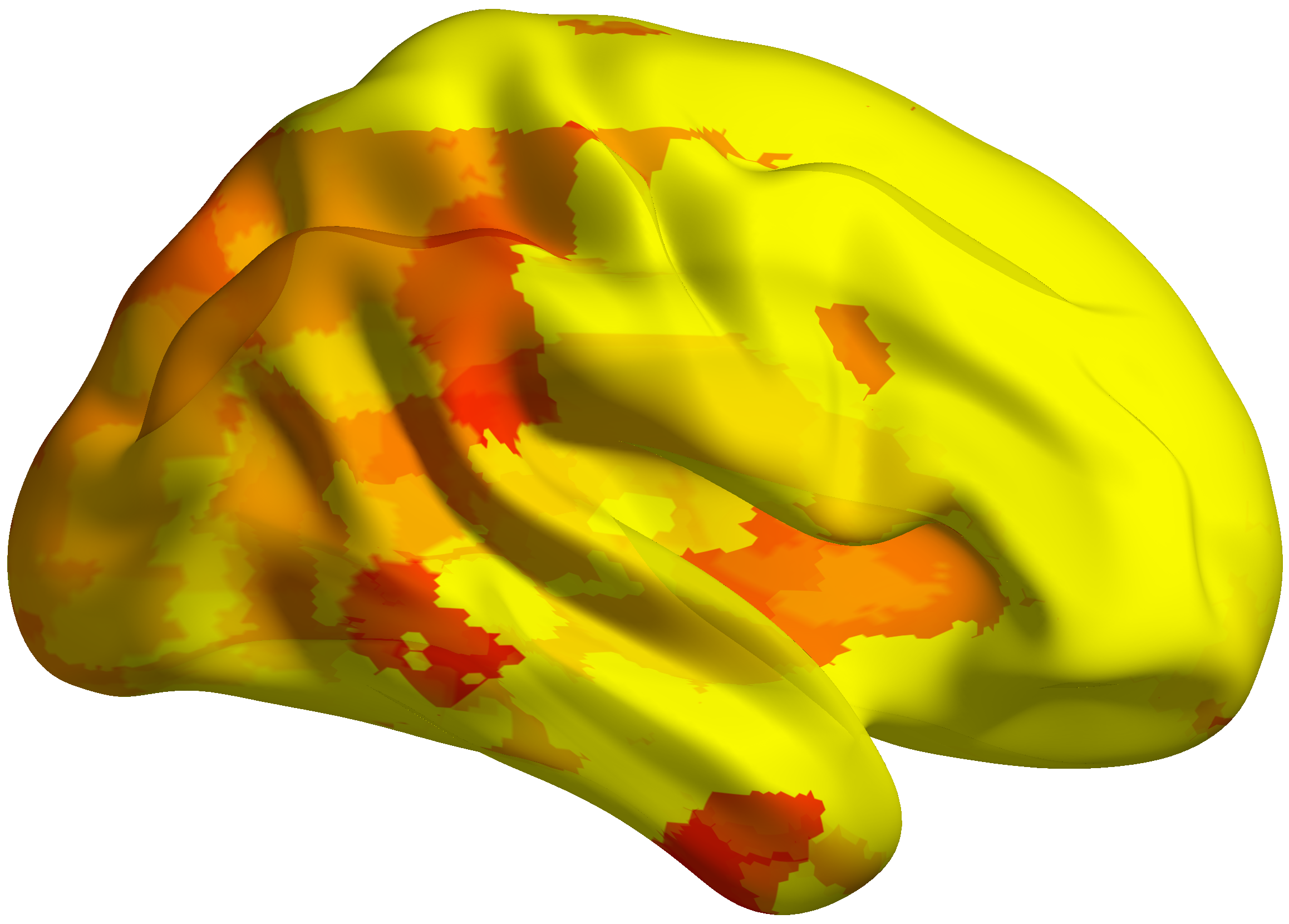} }}
    \hspace{1ex}~\subfloat[][\centering Up View]{{\includegraphics[width=0.24\textwidth]{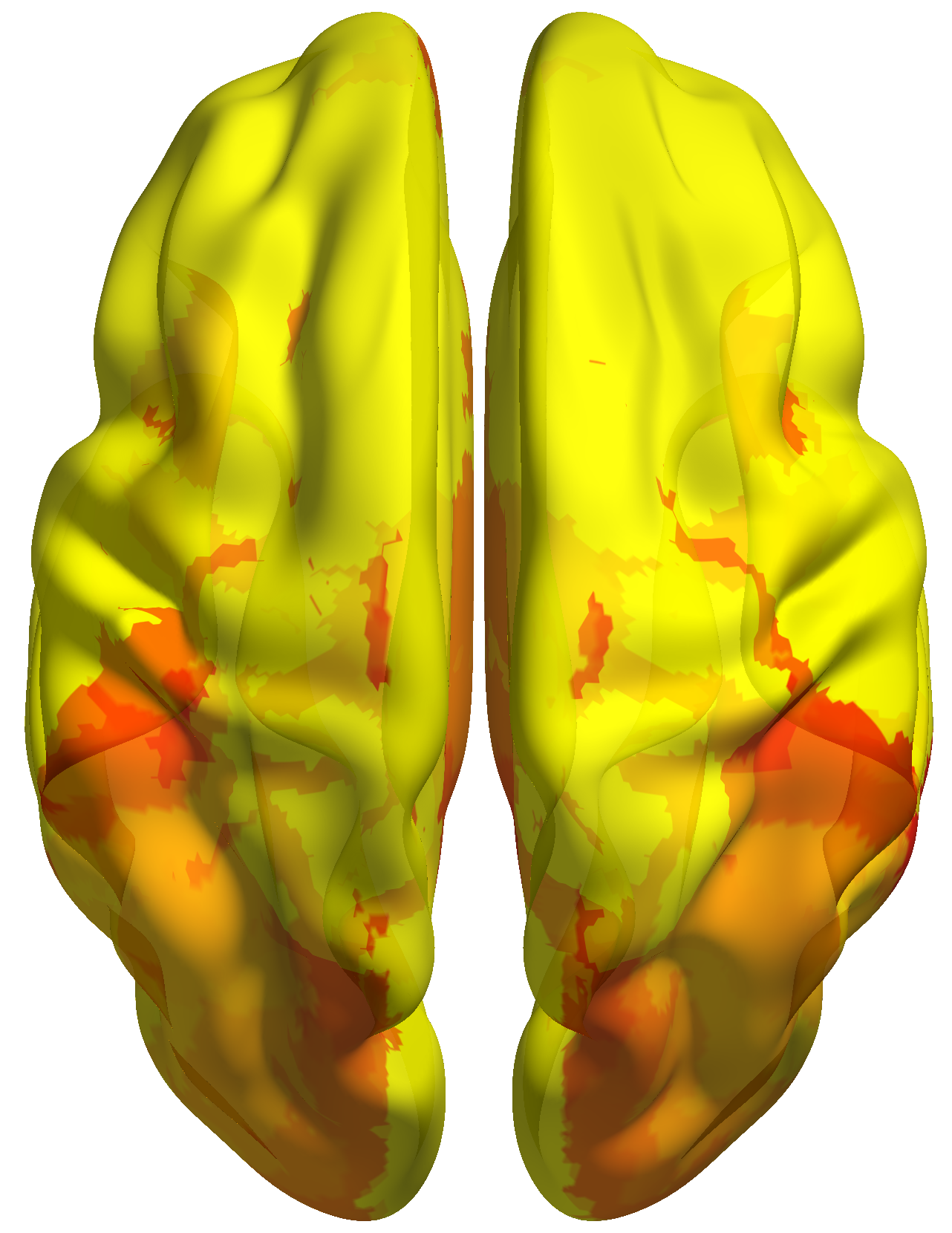} }}
    \caption{Distribution of anomalous edges in ADHD group.}
    \label{fig:adhd_dist}
\end{figure}
\subsection{Additional Results on Brain Networks} In this experiment, we investigate how anomalous edges found by \model{} are distributed in the brain.  
\autoref{fig:adhd_dist} reports the average distribution of anomalous edges in the brain networks of people living with ADHD. Most anomalous edges found by \model{} have a vertex in the Occipital lobes. Moreover, the Temporal lobes are the brain regions with the most anomalous connections with the Occipital lobes. These findings can help to reveal new insights into understanding ADHD and the regions of the brain that are connected to its symptoms. These findings also show the potential of \model{} to extract features that can help predict brain diseases or disorders. 

\section{Reproducibility}
\label{app:code}
The code, datasets, and supplements are available at \url{https://github.com/ubc-systopia/ANOMULY}.

\end{document}